\documentclass[conference]{IEEEtran}
\IEEEoverridecommandlockouts

\usepackage{cite}
\usepackage{amsmath,amssymb,amsfonts}
\usepackage{algorithmic}
\usepackage{algorithm}
\usepackage{graphicx}
\usepackage{hyperref}
\usepackage{textcomp}
\usepackage{xcolor}
\usepackage{url}
\usepackage{booktabs}
\usepackage{multirow}
\usepackage{siunitx}
\def\BibTeX{{\rm B\kern-.05em{\sc i\kern-.025em b}\kern-.08em
    T\kern-.1667em\lower.7ex\hbox{E}\kern-.125emX}}

\begin{document}

\title{Graph-Conditioned Mixture of Graph Neural Network Experts for Traffic Forecasting
\thanks{This work was supported by the Emerging Projects program, Infotech Oulu; European Commission (101137711); European Regional Development Fund
(A81373,A81376,A81568,A91867); Research Council of Finland (323630); the Strategic Research Council affiliated
with Academy of Finland (372355) and Business Finland
(8754/31/2022).}
\thanks{\textcopyright~2026 IEEE. Personal use of this material is permitted. Permission from IEEE must be obtained for all other uses, in any current or future media, including reprinting/republishing this material for advertising or promotional purposes, creating new collective works, for resale or redistribution to servers or lists, or reuse of any copyrighted component of this work in other works.}

}

\author{\IEEEauthorblockN{Amirhossein Ghaffari, Saeid Sheikhi, Ekaterina Gilman}
\IEEEauthorblockA{Future Computing Group, University of Oulu\\
Oulu, Finland\\
Email: firstname.lastname@oulu.fi}}

\maketitle

\begin{abstract}
Spatio-temporal forecasting on sensor graphs is commonly tackled with a single backbone architecture applied uniformly across all nodes, although graph regions can exhibit different dynamics. Road segments differ in functional class, structure, and traffic behavior, suggesting that node-wise expert specialization can be useful. We propose \emph{GC-MoE}, a graph-conditioned mixture of experts framework that assigns each node a personalized combination of frozen forecasting experts based on graph topology and the recent traffic input window. GC-MoE combines frozen pretrained spatio-temporal GNN experts with an input-aware, spatially contextualized router while training only a lightweight routing module. We also study a bounded graph-conditioned output refinement layer as an optional extension and include node-adaptive ST-LoRA adapters only as an ablation diagnostic. Across four standard benchmarks (PEMS04, PEMS07, METR-LA, and PEMS-BAY), GC-MoE improves MAE over a zero-parameter ensemble baseline, with competitive RMSE and MAPE, while training only ${\sim}$17K parameters on top of 1.5M frozen expert weights. The implementation is available at \href{https://github.com/Ahghaffari/gc_moe}{\url{https://github.com/Ahghaffari/gc_moe}}.
\end{abstract}

\begin{IEEEkeywords}
Spatio-temporal forecasting, traffic prediction, mixture of experts, graph neural networks, fine-tuning.
\end{IEEEkeywords}

\section{Introduction}
Spatio-temporal (ST) forecasting underpins critical urban analytics tasks, such as traffic speed/flow prediction, in which measurements arrive from sensors connected by a road network and evolve over time. Graph neural network (GNN) backbones that couple spatial structure with temporal dynamics, including diffusion recurrent models, graph-convolutional architectures, and spectral variants, are strong baselines for these problems \cite{li2017diffusion,yu2017spatio,wu2019graph,cao2020spectral}. Recent work has also emphasized the importance of the full spatio-temporal prediction pipeline, from spatial mapping and graph construction to model training and evaluation, highlighting the impact of graph design and preprocessing choices on downstream forecasting performance \cite{ghaffari2025stm}. Despite steady progress in model design, an important practical limitation remains: Different parts of the network may exhibit distinct dynamics due to differences in topology, road function, and connectivity, suggesting that a uniform backbone may be suboptimal.

At the same time, the research progress in spatio-temporal graph neural network (ST-GNN) backbones suggests complementary strengths. Diffusion-based models capture multi-hop propagation~\cite{li2017diffusion}; spectral graph convolutions capture smooth graph signals~\cite{yu2017spatio}; adaptive-graph models learn node-specific structure~\cite{bai2020agcrn}. A natural solution is to combine multiple architectures. Classic ensembling (e.g., uniform averaging) improves robustness~\cite{dietterich2000ensemble}, but ignores that the best expert may vary by node and condition. Learned ensembling via a meta-learner can improve combinations~\cite{jacobs1991adaptive}, yet typical routers depend primarily on input features and do not explicitly encode \emph{graph-topological descriptors} of nodes~\cite{shazeer2017outrageously,fedus2022switch} or exploit \emph{spatial neighbor context} to detect network-wide congestion propagation.

Parameter-efficient fine-tuning (PEFT) offers another approach that freezes a backbone and trains only small adapter modules, such as low-rank adaptation (LoRA) \cite{hu2022lora}. Recent ST-LoRA variants adapt spatio-temporal forecasting using node-adaptive low-rank modules with small trainable budgets \cite{ruan2025st}. However, PEFT alone does not address architectural heterogeneity; even a well-adapted single expert may remain suboptimal for certain node roles. Moreover, in a multi-expert setting, the interaction between routing and adapter corrections remains poorly understood.

We introduce \emph{GC-MoE} (\underline{G}raph \underline{C}onditioned \underline{M}ixture \underline{o}f \underline{E}xperts routing for ST-GNN forecasting), a modular framework that (i) pretrains multiple diverse ST-GNN experts, (ii) freezes them as an expert set, (iii) learns an \emph{input-aware, spatially contextualized routing mechanism} that assigns per-node expert weights using both static topology features and a dynamic pathway driven by temporally attended input signals with spatial message passing. We additionally study a lightweight graph-conditioned \emph{output refinement} layer as an optional extension. We also evaluate node-adaptive ST-LoRA adapters \cite{hu2022lora,ruan2025st} as an optional add-on and report the results.

The main contributions of this work are:
\begin{itemize}
  \item \emph{Input-aware, spatially contextualized graph-conditioned routing.} We propose a dual-pathway router that fuses static topology descriptors with a dynamic representation computed via temporal attention over the input window and spatial neighbor message passing, enabling expert selection that adapts to current traffic conditions, not just static topology.
  \item \emph{Frozen multi-architecture specialization with low trainable budget.} We combine diverse frozen pretrained experts through learned routing, training only \(\sim\)17K parameters while leveraging the representational capacity of the frozen expert set.
  \item \emph{Optional lightweight output refinement.} We study a bounded graph-conditioned refinement layer that can further improve performance in some settings at negligible parameter cost.
  \item \emph{Ablation analysis of lightweight extensions.} We evaluate the optional refinement module and use node-adaptive ST-LoRA adapters as a diagnostic ablation to study whether adapter-based expert modification complements routing.
\end{itemize}
The remainder of this paper is organized as follows. Section~\ref{sec:related} reviews related work and Section~\ref{sec:problem} formalizes the problem setup. Section~\ref{sec:method} presents the proposed GC-MoE framework in detail. Section~\ref{sec:experiments} describes the experimental setup, Section~\ref{sec:results} reports the empirical results, and Section~\ref{sec:discussion} discusses their implications, limitations, and relation to prior work. Finally, Section~\ref{sec:conclusion} concludes the paper and outlines future directions.

\section{Related Work}\label{sec:related}

\subsection{Spatio-Temporal Graph Forecasting}

ST-GNNs jointly model spatial dependencies among sensors and temporal dynamics for tasks such as traffic speed and flow prediction. Foundational approaches include DCRNN~\cite{li2017diffusion}, which couples diffusion convolutions with gated recurrent units; STGCN~\cite{yu2017spatio}, which replaces recurrence with purely convolutional temporal blocks interleaved with graph convolutions. Moreover, Graph WaveNet~\cite{wu2019graph} introduces adaptive adjacency learning and dilated causal convolutions. Spectral variants such as StemGNN~\cite{cao2020spectral} jointly apply graph Fourier and discrete Fourier transforms. Adaptive graph models such as AGCRN~\cite{bai2020agcrn} learn node-specific recurrent dynamics via adaptive graph convolution. Recent work by Ghaffari et al.~\cite{ghaffari2025stm} has further emphasized the importance of the full spatio-temporal prediction pipeline, from spatial mapping and graph construction to model training and evaluation, showing that graph design and preprocessing choices significantly impact downstream forecasting performance.

Recent architectures further improve performance and scalability. PDFormer~\cite{jiang2023pdformer} models propagation delay patterns with delay-aware spatial attention; STAEformer~\cite{liu2023staeformer} shows that transformers with spatio-temporal adaptive embeddings can be highly competitive; BigST~\cite{han2024bigst} targets large-scale road networks with linear complexity; and UniST~\cite{yuan2024unist} studies prompt-based universal urban spatio-temporal prediction.

Despite this rapid progress, many strong ST-GNN and transformer-based forecasting models still apply a single trained backbone uniformly to every node, overlooking the heterogeneous dynamics arising from differences in topology, road function, and connectivity across the network. This work addresses this limitation by conditioning per-node expert routing on the graph structure and the recent traffic input window.

\subsection{Mixture of Experts and Routing}

MoE models combine the outputs of multiple specialist sub-networks through a learned gating function that produces data-dependent mixture weights. The foundational MoE framework was introduced by Jacobs et al.~\cite{jacobs1991adaptive}, where a gating network learns to partition the input space among experts. Shazeer et al.~\cite{shazeer2017outrageously} scaled this paradigm with a sparsely-gated MoE layer, demonstrating significant capacity gains in language modeling while relying on additional load-balancing losses to prevent expert collapse. Switch Transformers~\cite{fedus2022switch} simplified sparse routing to a top-1 expert selection per token, achieving efficient scaling to trillion-parameter models. Zhou et~al.~\cite{zhou2022expertchoice} inverted the routing direction with expert-choice selection, where each expert independently selects its top-$k$ inputs, achieving natural load balance without auxiliary losses.

Recent years have seen a surge of large-scale MoE designs that refine routing and expert specialization. Mixtral~\cite{jiang2024mixtral} demonstrates a practical open-weight sparse MoE architecture in which each token is routed to 2 of 8 experts, achieving performance competitive with much larger dense models while activating only a fraction of parameters per forward pass. DeepSeekMoE~\cite{dai2024deepseekmoe} shows that fine-grained expert segmentation combined with shared expert isolation improves expert specialization, enabling a more nuanced division of knowledge between experts. Branch-Train-MiX (BTX)~\cite{sukhbaatar2024btx} trains expert LLMs independently on different data domains and subsequently merges them into a unified MoE with lightweight routing, demonstrating that independently trained (or frozen) experts can be effectively combined, a paradigm conceptually close to our approach of routing over frozen pretrained backbones. On the routing-mechanism front, Puigcerver et al.~\cite{puigcerver2024softmoe} propose Soft~MoE, which replaces discrete token-to-expert assignment with a fully differentiable soft assignment via learned slot projections, avoiding the load-balancing and training instability issues inherent in complex sparse routing.

In the spatio-temporal domain, TESTAM~\cite{lee2024testam} is the work most closely related to our work. It introduces a time-enhanced spatio-temporal attention model with a mixture of experts for traffic forecasting, where different experts specialize in different temporal traffic patterns (e.g., recurring vs.\ non-recurring congestion). However, TESTAM differs from GC-MoE in several key aspects. TESTAM is an end-to-end trained MoE architecture designed to model different temporal and spatio-temporal traffic patterns, including recurring and non-recurring regimes. In contrast, GC-MoE studies a frozen-expert regime in which independently pretrained and architecturally heterogeneous ST-GNN backbones are kept fixed, and only a lightweight graph-conditioned router is trained. Moreover, GC-MoE explicitly conditions per-node routing on hand-crafted graph topology descriptors together with spatially propagated dynamic traffic context.

Compared with prior MoE-based traffic forecasting models, GC-MoE specifically focuses on graph-conditioned, per-node soft routing over heterogeneous frozen expert architectures, a setting that is distinct from end-to-end MoE training over jointly optimized experts. GC-MoE fills this gap with a dual-pathway router that fuses static topology features with a dynamic, spatially propagated traffic context representation to produce per-node expert mixture weights.

\subsection{Parameter-Efficient Fine-Tuning and LoRA}

PEFT methods adapt large pretrained models by updating only a small subset of parameters while keeping the backbone frozen. LoRA~\cite{hu2022lora} injects trainable low-rank residual matrices into frozen weight matrices, enabling adaptation without increasing inference latency. QLoRA~\cite{dettmers2023qlora} further reduces memory requirements by combining 4-bit quantization with LoRA adapters, enabling the fine-tuning of massive models on limited hardware. DoRA~\cite{liu2024dora} decomposes pretrained weights into magnitude and direction components and applies low-rank adaptation only to the directional component, improving learning capacity and stability over standard LoRA. In the spatio-temporal domain, ST-LoRA~\cite{ruan2025st} extends LoRA with node-adaptive low-rank modules that account for spatial heterogeneity across the sensor graph, achieving competitive forecasting performance with small trainable budgets. Budget allocation approaches such as AdaLoRA~\cite{zhang2023adalora} adaptively distribute rank budgets across weight matrices based on importance scores.

More recently, the interaction between PEFT adapters and MoE routing has received increasing attention. LoRAMoE~\cite{dou2024loramoe} investigates combining LoRA adapters with MoE-style routing in large language models and finds that naive integration can cause world-knowledge forgetting and training conflicts, requiring careful architectural design to preserve pretrained capabilities. This observation is directly relevant to GC-MoE, where routing is learned over frozen expert backbones. Motivated by this, we include node-adaptive ST-LoRA adapters as an ablation within our frozen multi-expert setting to test whether lightweight expert adaptation complements or interferes with graph-conditioned routing. The resulting findings are reported in Section~\ref{sec:results}.

\section{Problem Setup}
\label{sec:problem}
Let $\mathcal{G}=(\mathcal{V},\mathcal{E},A)$ be a weighted sensor graph with $|\mathcal{V}|=N$ nodes and adjacency matrix $A\in\mathbb{R}^{N\times N}$. We denote by $\tilde{A}$ a normalized version of $A$ used for one-hop message passing (e.g., row-normalized or symmetrically normalized with self-loops).

Given a historical input window
\[
X \in \mathbb{R}^{B \times T \times N \times D},
\]
where $B$ is the batch size, $T$ is the history length, $N$ is the number of nodes, and $D$ is the input feature dimension, the goal is to predict the next $H$ steps
\[
Y \in \mathbb{R}^{B \times H \times N \times D_{\text{out}}},
\]
where $H$ is the forecasting horizon and $D_{\text{out}}$ is the output feature dimension.

We construct a set of $E$ pretrained expert forecasters $\{f^{(e)}\}_{e=1}^{E}$, each producing an expert prediction
\[
\hat{Y}^{(e)} = f^{(e)}(X), \qquad \hat{Y}^{(e)}\in\mathbb{R}^{B\times H\times N\times D_{\text{out}}}.
\]
For each sample $b$ and node $n$, a router outputs mixture weights $w_n^{(b,e)}\ge 0$ such that $\sum_{e=1}^E w_n^{(b,e)}=1$. The mixture prediction is computed node-wise as
\begin{equation}
\hat{Y}_{b,:,n,:}^{\text{mix}}
=
\sum_{e=1}^{E}
w_{n}^{(b,e)}\,
\hat{Y}_{b,:,n,:}^{(e)}.
\label{eq:moe_combination}
\end{equation}

\section{Methodology}
\label{sec:method}
\begin{figure*}[t]
    \centering
    \includegraphics[width=1.0\linewidth]{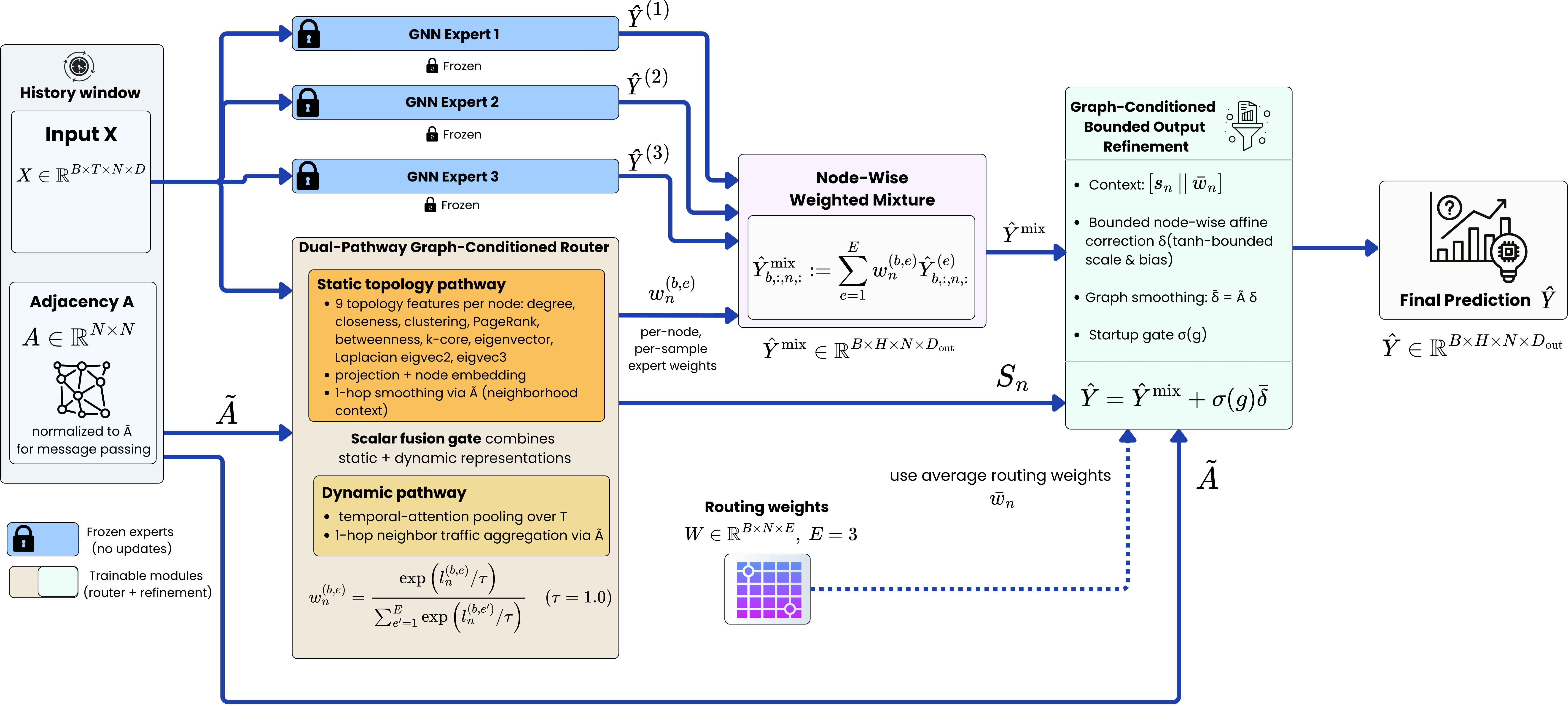}
    \caption{Overview of the GC-MoE design space. Given a historical window $X \in \mathbb{R}^{B \times T \times N \times D}$ and adjacency $A \in \mathbb{R}^{N \times N}$, $E$ frozen spatio-temporal GNN experts produce $\{\hat{Y}^{(e)}\}_{e=1}^{E}$. A dual-pathway graph-conditioned router combines static topology descriptors $\mathbf{s}_n$ with an input-aware dynamic representation to output per-node mixture weights $w_n^{(b,e)}$. The core GC-MoE model uses the router-based node-wise mixture $\hat{Y}^{\text{mix}}$ as its prediction. An optional bounded graph-conditioned refinement module, shown on the right, can be added as a lightweight correction layer and is evaluated separately in ablation.}
    \label{fig:architecture}
\end{figure*}

GC-MoE consists of two core components: (i) frozen pretrained experts and (ii) a dual-pathway graph-conditioned router. In addition, we study a lightweight graph-conditioned output refinement module as an optional add-on, and node-adaptive ST-LoRA adapters as an ablation-only extension. Figure~\ref{fig:architecture} illustrates the full design space; the core GC-MoE model used in the main results includes only frozen experts and the graph-conditioned router, while refinement is evaluated separately in ablation. The framework supports an arbitrary number of experts and is not tied to the specific three backbones used in our experiments. We freeze multiple pretrained spatio-temporal GNN experts and learn a graph-conditioned router that outputs per-node mixture weights from static topology and dynamic traffic context. Optionally, a bounded graph-conditioned refinement module can be added on top of the mixed prediction as a lightweight correction layer.

\subsection{Frozen Expert Backbones}
\label{subsec:experts}
We use $E=3$ diverse ST-GNN backbones, namely STGCN~\cite{yu2017spatio}, GWNet~\cite{wu2019graph}, AGCRN~\cite{bai2020agcrn}. Each expert is pretrained to converge on the target dataset and then frozen during MoE training. Expert parameters are not updated; only routing and refinement parameters are trained.

\subsection{Dual-Pathway Graph-Conditioned Router}
\label{subsec:router}

\paragraph{Static topology descriptor}
For each node $n$, we compute a normalized topology vector $\mathbf{s}_n \in \mathbb{R}^{d_s}$ with $d_s=9$:

\begin{equation}
\begin{aligned}
\mathbf{s}_n = \big[&\deg, \mathrm{close}, \mathrm{clust}, \mathrm{PR}, \mathrm{btw}, \\
&\mathrm{kcore}, \mathrm{eig}, u_2, u_3\big]_n \in [0,1]^{9}
\end{aligned}
\end{equation}

where the entries correspond to (normalized) degree, closeness centrality, clustering coefficient, PageRank, betweenness centrality, k-core number, eigenvector centrality, and the node-wise entries of the second (Fiedler) and third Laplacian eigenvectors. The 9 topology features were chosen to cover local connectivity, centrality, and spectral position.

\paragraph{Static representation with neighborhood smoothing}
Let $\mathrm{Proj}(\cdot)$ be a learnable linear map to $\mathbb{R}^{d_r}$ and let $\mathbf{e}_n\in\mathbb{R}^{d_r}$ be a learnable node embedding. The initial static embedding is
\begin{equation}
\mathbf{r}_n^{\text{static}} =
\mathrm{Proj}(\mathbf{s}_n) + \mathbf{e}_n,
\qquad \mathbf{r}_n^{\text{static}}\in\mathbb{R}^{d_r}.
\end{equation}
Because both $\mathrm{Proj}(\cdot)$ and $\mathbf{e}_n$ are learnable, the static pathway remains trainable even though it is built from fixed topology descriptors. We then apply one-hop neighborhood smoothing with a learnable scalar gate $\gamma\in(0,1)$:
\begin{equation}
\mathbf{r}_n^{\text{static}} \leftarrow
\mathbf{r}_n^{\text{static}} +
\gamma \,(\tilde{A}\mathbf{r}^{\text{static}})_n.
\label{eq:static_smooth}
\end{equation}

The raw topology descriptors are fixed node-wise summaries. Although several encode global graph position, they do not directly aggregate the structural context of neighboring nodes. We therefore apply one-hop smoothing to contextualize each node’s representation. Although the topology descriptors and adjacency are fixed, the projection, node embeddings, and smoothing gate are learnable; therefore, the smoothed static representation is recomputed during each forward pass using the current parameters. In practice, this branch is lightweight and adds negligible cost relative to expert inference.

Concretely, $(\tilde{A}\mathbf{r}^{\text{static}})_n = \sum_{j\in\mathcal{N}(n)}\tilde{A}_{nj}\,\mathbf{r}_j^{\text{static}}$ aggregates the embeddings of all nodes adjacent to $n$, weighted by normalized edge strengths.
The update in \eqref{eq:static_smooth} then blends this neighborhood summary back into $\mathbf{r}_n^{\text{static}}$ with a learnable scalar gate $\gamma = \sigma(\hat{\gamma})$, where $\hat{\gamma}$ is an unconstrained parameter initialized at $0$ (so $\gamma\approx 0.5$ at the start of training).

The gate $\gamma$ controls the trade-off between \emph{self-reliance} (routing based solely on the node's own topology) and \emph{contextual awareness} (routing informed by the structural character of the surrounding subgraph). If neighbors of $n$ are predominantly high-centrality bottleneck nodes, the smoothed embedding shifts toward that regime, causing the router to treat $n$ as part of a critical corridor rather than an isolated hub.
Conversely, if $n$ is a high-degree node surrounded by peripheral sensors, the neighborhood signal pulls the embedding away from the hub prototype, enabling finer-grained expert specialization.

\paragraph{Dynamic pathway (temporal attention + spatial propagation)}
For each sample $b$ and node $n$, we compute a temporally attended dynamic representation:
\begin{equation}
\mathbf{h}_n^{(b)} =
\sum_{t=1}^{T}
\alpha_t^{(b,n)}\, g(X_{b,t,n,:}),
\qquad
\sum_{t=1}^{T}\alpha_t^{(b,n)}=1,
\label{eq:dynamic_temporal}
\end{equation}
where $g(\cdot)$ is a learnable feature map to $\mathbb{R}^{d_r}$ and $\alpha_t^{(b,n)}$ are attention weights over the $T$ history steps. We then inject the 1-hop neighbor context:
\begin{equation}
\mathbf{h}_n^{(b)} \leftarrow
\mathbf{h}_n^{(b)} + (\tilde{A}\mathbf{h}^{(b)})_n,
\label{eq:dynamic_spatial}
\end{equation}
where $\mathbf{h}^{(b)}\in\mathbb{R}^{N\times d_r}$ stacks $\{\mathbf{h}_n^{(b)}\}_{n=1}^{N}$.

\paragraph{Scalar fusion gate}
The router fuses static and dynamic representations using a scalar gate $\lambda_n^{(b)}\in(0,1)$:
\begin{equation}
\lambda_n^{(b)} =
\sigma\!\Big(\mathrm{MLP}\big([\mathbf{r}_n^{\text{static}} \,\|\, \mathbf{h}_n^{(b)}]\big)\Big),
\label{eq:fusion_gate_scalar}
\end{equation}
Here $\sigma(\cdot)$ denotes the logistic sigmoid. Then, the fused router embedding
\begin{equation}
\mathbf{r}_n^{(b)} =
\lambda_n^{(b)}\, \mathbf{r}_n^{\text{static}}
+
\big(1-\lambda_n^{(b)}\big)\, \mathbf{h}_n^{(b)}.
\label{eq:fused_repr}
\end{equation}

Both $\mathbf{r}_n^{\text{static}}$ and $\mathbf{h}_n^{(b)}$ lie in $\mathbb{R}^{d_r}$, hence the fused representation $\mathbf{r}_n^{(b)}\in\mathbb{R}^{d_r}$.

\paragraph{Routing logits and weights}
The dynamic routing head outputs logits
\begin{equation}
\mathbf{l}_n^{(b,\text{dyn})}
=
W_2\,\mathrm{ReLU}(W_1 \mathbf{r}_n^{(b)}),
\qquad
\mathbf{l}_n^{(b,\text{dyn})}\in\mathbb{R}^{E}.
\label{eq:dynamic_logits}
\end{equation}
Here $\mathrm{ReLU}(x)=\max(x,0)$, $W_1\in\mathbb{R}^{d_h\times d_r}$ and $W_2\in\mathbb{R}^{E\times d_h}$ are learnable matrices (with hidden size $d_h$), and $\mathbf{l}_n^{(b,\text{dyn})}$ denotes pre-softmax routing logits. The final logits include a learnable per-node routing bias table $\boldsymbol{\theta}_n\in\mathbb{R}^{E}$:
\begin{equation}
\mathbf{l}_n^{(b)} =
\boldsymbol{\theta}_n + \mathbf{l}_n^{(b,\text{dyn})}.
\label{eq:final_logits}
\end{equation}
Let $l_n^{(b,e)}$ denote the $e$-th component of $\mathbf{l}_n^{(b)}$; we use a fixed softmax temperature $\tau=1.0$:
\begin{equation}
w_{n}^{(b,e)} =
\frac{\exp(l_{n}^{(b,e)}/\tau)}
{\sum_{e'=1}^{E}\exp(l_{n}^{(b,e')}/\tau)},
\qquad \tau=1.0.
\label{eq:router_softmax}
\end{equation}

\paragraph{Load-balancing loss}
To discourage router collapse, we use the standard load-balancing objective:
\begin{equation}
\mathcal{L}_{\text{balance}} =
E \sum_{e=1}^{E} f_e\, P_e,
\label{eq:balance}
\end{equation}
where, averaged over the batch,
\begin{equation}
\begin{aligned}
f_e &= \frac{1}{BN}\sum_{b=1}^{B}\sum_{n=1}^{N}\mathbf{1}\!\left[\arg\max_{e'} w_{n}^{(b,e')} = e\right], \\
P_e &= \frac{1}{BN}\sum_{b=1}^{B}\sum_{n=1}^{N} w_{n}^{(b,e)}.
\end{aligned}
\label{eq:balance_terms}
\end{equation}

\subsection{Optional Graph-Conditioned Bounded Output Refinement}
\label{subsec:refine}
Let $\hat{Y}^{\text{mix}}$ denote the mixture prediction from \eqref{eq:moe_combination}. As an optional extension, we consider a lightweight refinement module that predicts a bounded, graph-smoothed affine correction on top of the router-based mixture output. For each node $n$, we form a context vector
\begin{equation}
\mathbf{c}_n = [\mathbf{s}_n \,\|\, \bar{\mathbf{w}}_n]\in\mathbb{R}^{d_s+E},
\label{eq:refine_context}
\end{equation}
where $\bar{\mathbf{w}}_n\in\mathbb{R}^{E}$ is the average routing weight vector for node $n$ across the current batch,
i.e., $\bar{\mathbf{w}}_n=\frac{1}{B}\sum_{b=1}^{B}\mathbf{w}_n^{(b)}$ and $\mathbf{w}_n^{(b)}=[w_n^{(b,1)},\dots,w_n^{(b,E)}]$.

Two small networks produce per-node scale and bias (both bounded via $\tanh$):
\begin{align}
m_n &= \beta \,\tanh(\text{scale\_net}(\mathbf{c}_n)), \\
b_n &= \beta \,\tanh(\text{bias\_net}(\mathbf{c}_n)),
\end{align}
where $m_n,b_n\in(-\beta,\beta)$ are broadcast across $(H,D_{\text{out}})$ for node $n$ and $\beta \in (0, 1)$ is a hyperparameter bounding the maximum refinement. In our experiments, we set $\beta = 0.3$. The bounded affine correction is
\begin{equation}
\delta = \hat{Y}^{\text{mix}} \odot m + b,
\label{eq:delta_affine}
\end{equation}
where $m$ and $b$ denote the node-wise scale and bias tensors broadcast to the shape of $\hat{Y}^{\text{mix}}$.
We then apply one-hop graph smoothing to the correction:
\begin{equation}
\bar{\delta} = \tilde{A}\,\delta.
\label{eq:delta_smooth}
\end{equation}
Finally, the refined prediction is
\begin{equation}
\hat{Y}
=
\hat{Y}^{\text{mix}}
+
\sigma(g)\,\bar{\delta},
\label{eq:refine_final}
\end{equation}
where $g$ is a learnable startup gate initialized to a negative value so that $\sigma(g)\approx 0$ at early training stages.

\subsection{Optional: Per-Expert Node-Adaptive ST-LoRA Adapters (Ablation Only)}
\label{subsec:stlora}
We evaluate node-adaptive ST-LoRA adapters inspired by LoRA \cite{hu2022lora} and ST-LoRA \cite{ruan2025st} as an optional add-on. In our experiments, adapters \emph{degraded} performance in the routed multi-expert setting (Section~\ref{sec:results}), suggesting that adapter corrections can interfere with routing. We therefore do not include adapters in the final model.

\subsection{Training Objective}
The total objective is
\begin{equation}
\mathcal{L}
=
\mathcal{L}_{\text{MAE}}
+
\lambda_1 \mathcal{L}_{\text{balance}}
+
\lambda_2 \mathcal{L}_{\text{entropy}}.
\label{eq:loss_total}
\end{equation}
We use masked MAE:
\begin{equation}
\mathcal{L}_{\text{MAE}} =
\frac{1}{|\Omega|}
\sum_{(b,h,n)\in\Omega}
\left\|Y_{b,h,n,:}-\hat{Y}_{b,h,n,:}\right\|_1,
\label{eq:mae}
\end{equation}
where $\Omega$ indexes non-missing targets. The entropy term encourages confident (peaked) routing:
\begin{equation}
\mathcal{L}_{\text{entropy}}
=
\frac{1}{BN}
\sum_{b=1}^{B}\sum_{n=1}^{N}
\left(
-\sum_{e=1}^{E}
w_{n}^{(b,e)}
\log w_{n}^{(b,e)}
\right).
\label{eq:entropy}
\end{equation}
The entropy term mildly encourages sharper per-node expert preferences, while the load-balancing term prevents collapse to a single expert at the population level. In practice, we found that the combination stabilized training and yielded selective but non-degenerate routing. We selected $\lambda_1=0.01$ and $\lambda_2=0.5$ based on validation performance and stable expert utilization.

\subsection{Trainable Parameters}
In the core GC-MoE model, only the router parameters (including the fusion-gate MLP and the per-node bias table $\{\boldsymbol{\theta}_n\}$), node embeddings $\{\mathbf{e}_n\}$, and related routing parameters are trainable (approximately $17$K parameters). Expert backbones are frozen and are not updated during training. When the optional refinement module is enabled, the trainable parameter count increases slightly to approximately $18$K.

\section{Experiments}
\label{sec:experiments}

\subsection{Datasets and Setup}

We evaluate the GC-MoE framework on four widely used traffic forecasting benchmarks summarized in Table~\ref{tab:datasets}. \textbf{PEMS04} and \textbf{PEMS07} are traffic flow datasets collected by the California Department of Transportation (Caltrans) Performance Measurement System~\cite{guo2019attention}, aggregated at 5-minute intervals. \textbf{METR-LA} and \textbf{PEMS-BAY} are traffic speed datasets collected from loop detectors in Los Angeles County and the San Francisco Bay Area~\cite{li2017diffusion}, respectively, also at 5-minute resolution. For all datasets, the input features consist of the traffic measurement and a time-of-day indicator, and the output is a single-channel forecast. We use the standard chronological splits: 6:2:2 (train/val/test) for PEMS04 and PEMS07, and 7:1:2 for METR-LA and PEMS-BAY, following~\cite{li2017diffusion,wu2019graph}.

\begin{table}[t]
\centering
\caption{Dataset statistics. All datasets use 5-minute aggregation intervals}
\label{tab:datasets}
\begin{tabular}{lcccc}
\toprule
Dataset & Nodes & Edges & Time Steps & Duration \\
\midrule
PEMS04   & 307  & 340   & 16{,}992 & 59 days  \\
PEMS07   & 883  & 866 & 28{,}224 & 98 days  \\
METR-LA  & 207  & 1{,}515 & 34{,}272 & 119 days \\
PEMS-BAY & 325  & 2{,}369 & 52{,}116 & 181 days \\
\bottomrule
\end{tabular}
\end{table}

Following common practice~\cite{li2017diffusion,wu2019graph,yu2017spatio}, we use a history window of $T=12$ steps to predict the next $H=12$ steps.
Adjacency matrices are constructed from road-network distances as provided by the benchmark datasets and normalized using the double-transition scheme~\cite{wu2019graph}. We report three standard metrics: Mean Absolute Error (MAE), Root Mean Squared Error (RMSE), and Mean Absolute Percentage Error (MAPE), all averaged over the 12-step horizon (lower is better for all metrics).

\subsection{Expert Backbones}
We instantiate $E=3$ diverse frozen experts, each representing a different family of ST-GNN architectures:
\begin{itemize}
    \item \textbf{Graph WaveNet (GWNet)}~\cite{wu2019graph}: diffusion convolutions with adaptive adjacency learning and dilated causal temporal convolutions (${\sim}$303K parameters on PEMS04).
    \item \textbf{STGCN}~\cite{yu2017spatio}: spectral graph convolutions (Chebyshev polynomials) interleaved with 1-D convolutional temporal blocks (${\sim}$298K parameters on PEMS04).
    \item \textbf{AGCRN}~\cite{bai2020agcrn}: adaptive graph convolution with node-specific GRU-based recurrent dynamics (${\sim}$903K parameters on PEMS04).
\end{itemize}
Each expert is independently pretrained to convergence on the target dataset using Adam~\cite{kingma2014adam} with a learning rate of $10^{-3}$, weight decay of $5\times 10^{-4}$, batch size 64, and early stopping with patience 15 (maximum 200 epochs).
After pretraining, all expert parameters are frozen and are not updated during MoE training.

\subsection{GC-MoE Training}
During MoE training, only the router and (optionally) the output refinement module are optimized using Adam with a learning rate of $10^{-3}$, weight decay of $10^{-5}$, gradient clipping at norm 5, batch size 64, and early stopping with patience 15. The router embedding dimension is $32$, and the routing MLP hidden size is $32$. We use a fixed softmax temperature $\tau=1.0$. The output refinement hidden dimension is 32 with a bound of $\pm 0.3$. Training is conducted on Tesla P100 16\,GB GPUs.

\subsection{Baselines and Ablations}
We compare GC-MoE (frozen experts + graph-conditioned router) against the following configurations:
\begin{itemize}
    \item \textbf{Single experts}: Each frozen expert evaluated independently (GWNet, STGCN, AGCRN).
    \item \textbf{Ens-Avg}: Uniform averaging of the three frozen expert outputs with zero learned parameters, serving as a strong non-parametric baseline.
    \item \textbf{Component ablations} (on PEMS04): Router only (core GC-MoE), router + adapters, router + refinement, router + refinement + adapters, and MoE partial fine-tune (experts and router unfrozen; LoRA base weights inside adapter blocks remain frozen by design).
    \item \textbf{Router ablations} (on PEMS04): Dense MLP (no graph features), Switch-style top-1 routing~\cite{fedus2022switch}, expert-choice routing~\cite{zhou2022expertchoice}, hash routing (parameter-free), and the proposed graph-conditioned router.
\end{itemize}

Ablation studies are conducted on PEMS04, which we use as a representative benchmark for controlled analysis because it has a standard scale for traffic forecasting experiments while keeping the computational cost of testing multiple routing and adaptation variants manageable. The final GC-MoE core model is then evaluated on all four benchmarks.

\section{Results}
\label{sec:results}

\subsection{Main Forecasting Performance}

\begin{table*}[t]
\caption{Performance comparison across four traffic forecasting benchmarks (lower is better). Bold denotes the best result for each metric and dataset. Ens-Avg is a uniform average of three frozen experts (zero learned parameters). GC-MoE uses the graph-conditioned router with frozen experts (${\sim}$17\,K trainable parameters)}
\label{tab:main}
\centering
\resizebox{\textwidth}{!}{
\begin{tabular}{lcccccccccccc}
\toprule
\multirow{2}{*}{Method}
  & \multicolumn{3}{c}{PEMS04}
  & \multicolumn{3}{c}{METR-LA}
  & \multicolumn{3}{c}{PEMS-BAY}
  & \multicolumn{3}{c}{PEMS07} \\
\cmidrule(lr){2-4}\cmidrule(lr){5-7}\cmidrule(lr){8-10}\cmidrule(lr){11-13}
 & MAE & RMSE & MAPE
 & MAE & RMSE & MAPE
 & MAE & RMSE & MAPE
 & MAE & RMSE & MAPE \\
\midrule
GWNet
  & 19.794 & 31.206 & 13.89\%
  & 3.103 & 6.146 & 8.54\%
  & 1.630 & 3.580 & 3.66\%
  & 21.984 & 35.291 & 9.71\% \\
STGCN
  & 20.508 & 32.176 & 14.54\%
  & 3.159 & 6.337 & 8.62\%
  & 1.676 & 3.678 & 3.88\%
  & 24.560 & 37.633 & 11.44\% \\
AGCRN
  & 18.902 & 30.428 & 13.53\%
  & 3.130 & 6.261 & 8.57\%
  & 1.633 & 3.578 & 3.73\%
  & 20.693 & 33.915 & 8.75\% \\
\midrule
Ens-Avg
  & 18.889 & \textbf{30.142} & \textbf{12.98\%}
  & 3.046 & \textbf{5.980} & 8.28\%
  & 1.590 & 3.474 & 3.61\%
  & 20.903 & 33.584 & 8.89\% \\
\textbf{GC-MoE (ours)}
  & \textbf{18.741} & 30.236 & \textbf{12.98\%}
  & \textbf{3.043} & 6.007 & \textbf{8.27\%}
  & \textbf{1.577} & \textbf{3.460} & \textbf{3.57\%}
  & \textbf{20.349} & \textbf{33.486} & \textbf{8.54\%} \\
\bottomrule
\end{tabular}}
\end{table*}

Table~\ref{tab:main} reports the main forecasting results across PEMS04, METR-LA, PEMS-BAY, and PEMS07. GC-MoE improves MAE over all single backbones and over the zero-parameter ensemble baseline across all four benchmarks, while RMSE and MAPE remain competitive and are best on several datasets. Improvements over Ens-Avg are modest but consistent, demonstrating that node-dependent, input-aware routing provides additional gains beyond simple averaging of diverse experts. The main comparison reports the core GC-MoE model for consistency across all four datasets; the optional refinement module was evaluated separately via an ablation study on PEMS04. On PEMS04, GC-MoE reduces MAE from 18.889 (Ens-Avg) to 18.741, and further to 18.723 with refinement. On METR-LA and PEMS-BAY, GC-MoE achieves the best MAE among all methods while maintaining competitive RMSE and MAPE.

\subsection{Component Ablation and Adapter Analysis}

\begin{table}[t]
\centering
\caption{Ablation study on PEMS04 (MAE; lower is better).
``Router only'' trains only the graph-conditioned router (${\sim}$17\,K params).
``+ Refinement'' adds the bounded output refinement layer.
``MoE partial fine-tune'' unfreezes the experts and the router but
keeps the LoRA base weights inside the ST-LoRA adapter blocks frozen
($1{,}563$K trainable out of $1{,}813$K total). ST-LoRA adapters
degrade performance}
\label{tab:ablation}
\begin{tabular}{lcc}
\toprule
Variant & Trainable Params & PEMS04 MAE \\
\midrule
Ens-Avg (baseline) & $0$ & 18.889 \\
Router only & ${\sim}17$K & 18.741 \\
Router + adapters & ${\sim}60$K & 19.681 \\
Router + refinement & ${\sim}18$K & \textbf{18.723} \\
Router + refine + adapters & ${\sim}61$K & 20.397 \\
MoE partial fine-tune & ${\sim}1{,}563$K & 18.757 \\
\bottomrule
\end{tabular}
\end{table}

Table~\ref{tab:ablation} presents the contribution of each component on PEMS04.
Two consistent patterns emerge:

\subsubsection{Router effectiveness} Training only the graph-conditioned router already improves over Ens-Avg, confirming that topology- and context-aware weighting is beneficial even without refinement.

\subsubsection{Output refinement gains} Adding the bounded graph-conditioned refinement layer yields an additional improvement (18.723 MAE), demonstrating that lightweight affine correction with graph smoothing can further reduce residual errors.

\subsubsection{Negative result on ST-LoRA adapters} Introducing node-adaptive ST-LoRA adapters degrades performance when combined with routing and refinement. This suggests a conflict between adapter-based expert modification and routing-based expert specialization. Consequently, adapters are excluded from the final model.

Therefore, in our PEMS04 experiments, MoE partial fine-tune did not outperform the lightweight routed configuration, reinforcing the effectiveness of frozen-expert routing.

\subsection{Router Architecture Comparison}

\begin{table}[t]
\centering
\caption{Router architecture comparison on PEMS04. All variants use the same frozen experts
and output refinement; only the router differs.
Expert-choice routing diverged and is omitted}
\label{tab:router}
\begin{tabular}{lccc}
\toprule
Router Type & MAE & RMSE & MAPE \\
\midrule
Dense MLP (no graph) & 19.475 & 30.562 & 14.59\% \\
Switch (sparse top-1) & 19.246 & 30.607 & 13.21\% \\
Hash routing (no learning) & 19.768 & 31.276 & 13.75\% \\
\textbf{Graph-conditioned (ours)} & \textbf{18.723} & \textbf{30.156} & \textbf{12.99\%} \\
\bottomrule
\end{tabular}
\end{table}

Table~\ref{tab:router} compares alternative routing strategies on PEMS04. The proposed graph-conditioned router achieves the best MAE, RMSE, and MAPE among learned routing mechanisms. We evaluated five routing strategies; four are reported in Table~\ref{tab:router}, while expert-choice routing diverged during training and is omitted from the table.

\subsubsection{GC-MoE Router (Ours)}
Combines a static pathway, nine topological features smoothed over 1-hop neighborhoods plus a learnable per-node bias, with a dynamic pathway that applies temporal attention followed by adjacency-based spatial propagation. A scalar gate fuses both pathways before softmax routing.

\subsubsection{Dense MLP}
A two-layer MLP maps a learned node embedding to expert weights without any graph features, serving as a non-graph-aware baseline.

\subsubsection{Switch (top-1)}
Inspired by Switch Transformers~\cite{fedus2022switch}, this router selects exactly one expert per node via a hard top-1 with a straight-through estimator, using node embeddings as input.

\subsubsection{Expert Choice (top-$k$)}
Following the paradigm of~\cite{zhou2022expertchoice}, each expert independently scores all nodes and selects its top-$k$, allowing variable expert assignment per node.

\subsubsection{Hash}
A parameter-free deterministic assignment ($\text{node\_id} \bmod E$) that controls for whether any learned routing is beneficial.

Dense MLP routing (no graph features) performs worse, indicating that static topology descriptors and spatial propagation are important.
Sparse top-1 (Switch-style) routing underperforms soft routing, suggesting that soft mixtures better capture gradual specialization across nodes. Deterministic hash routing also underperforms, confirming that learned, topology-aware routing is necessary.

\subsection{Parameter Efficiency}

\begin{table}[t]
\centering
\caption{Parameter efficiency on PEMS04}
\label{tab:params}
\begin{tabular}{lcccc}
\toprule
Method & Total & Trainable & Train.\,\% & MAE \\
\midrule
GWNet (single) & 303K & 303K & 100\% & 19.794 \\
STGCN (single) & 298K & 298K & 100\% & 20.508 \\
AGCRN (single) & 903K & 903K & 100\% & 18.902 \\
Ens-Avg & 1{,}504K & $0$ & $0\%$ & 18.889 \\
GC-MoE & 1{,}521K & ${\sim}17$K & 1.1\% & 18.741 \\
GC-MoE + refine & 1{,}522K & ${\sim}18$K & 1.2\% & 18.723 \\
MoE partial fine-tune & 1{,}813K & ${\sim}1{,}563$K & 86.3\% & 18.757 \\
\bottomrule
\end{tabular}
\end{table}

Table~\ref{tab:params} compares parameter efficiency on PEMS04. GC-MoE trains only $\sim$17K parameters for routing, or $\sim$18K with refinement, while freezing approximately 1.5M expert parameters. It achieves the best MAE among the compared configurations, and MoE partial fine-tuning does not improve over GC-MoE with refinement. The efficiency claim refers to trainable parameters during adaptation, not to expert pretraining or inference cost; since GC-MoE evaluates all frozen experts, its inference cost is closer to an ensemble than to a single backbone.

\section{Discussion}
\label{sec:discussion}
The results suggest that the main benefit of GC-MoE does not come from increasing backbone capacity, but from selectively composing the complementary strengths of diverse frozen experts at the node level. Compared with uniform averaging, the gains are consistent across all four benchmark datasets. This indicates that the improvement is attributable to graph-conditioned, input-aware routing rather than to additional expert capacity.

Among the four benchmarks, on the largest PEMS07 dataset, the single frozen experts exhibit noticeably larger performance disparities than on the other datasets, where expert metrics are comparatively closer. In this regime of higher expert heterogeneity, GC-MoE demonstrates clearer gains over both the individual experts and Ens-Avg, suggesting that learned routing is particularly beneficial when expert strengths are less uniform and the gap between experts is larger. In contrast, when experts are already closely matched, as on some of the smaller benchmarks, the room for improvement beyond uniform averaging is naturally smaller.

From the perspective of related work, the results position GC-MoE between classical ensembling and fully trained MoE systems. Unlike standard ensembles~\cite{dietterich2000ensemble}, GC-MoE does not assign the same weights to all nodes, and unlike typical MoE formulations~\cite{jacobs1991adaptive,shazeer2017outrageously,fedus2022switch}, it operates over heterogeneous frozen spatio-temporal experts rather than jointly trained experts of a shared architecture. The strong performance of the proposed router relative to dense MLP, sparse top-1, and deterministic hash routing further suggests that explicit graph-topological conditioning and soft node-wise mixtures are important design choices in spatio-temporal forecasting.

The adapter ablation also provides an informative result. While parameter-efficient adaptation is often beneficial in single-model settings~\cite{hu2022lora,ruan2025st}, our results suggest that, in a routed multi-expert framework, node-adaptive adapter corrections may interfere with, rather than complement, expert specialization. This observation appears consistent with prior work showing that the design of LoRA- and MoE-style combinations can strongly affect performance~\cite{dou2024loramoe}, and suggests that parameter-efficient adaptation is not necessarily complementary to routing in our frozen multi-expert setting.

It is also worth noting that the present study has several limitations. First, the current expert pool contains only three backbones, all from the ST-GNN family; incorporating more diverse experts may reveal stronger specialization effects. Second, the static routing pathway relies on hand-crafted topology descriptors. Although these features are inexpensive and interpretable, they may not fully capture richer structural characteristics that could be learned through more expressive positional or structural encodings. Finally, our experiments are limited to traffic forecasting benchmarks. Although the proposed framework is general at the level of graph-conditioned expert routing, different application domains may exhibit different graph semantics, temporal dynamics, and informative structural descriptors, so the effectiveness of the current design beyond traffic data remains to be validated.

Overall, the findings support the view that lightweight graph-conditioned routing is a practical way to exploit complementary frozen spatio-temporal experts, especially when expert behaviors differ meaningfully, while also revealing important constraints on the interaction between routing and parameter-efficient adaptation.

\section{Conclusion and Future Work}
\label{sec:conclusion}

We presented GC-MoE, a graph-conditioned mixture of experts framework for spatio-temporal traffic forecasting that routes each node to a personalized combination of frozen pretrained ST-GNN experts.
The core of the approach is a dual-pathway router that fuses static topology descriptors, smoothed over one-hop neighborhoods, with a dynamic, temporally attended and spatially propagated representation of the current traffic state, producing per-node soft expert mixture weights. We additionally studied a lightweight bounded-output refinement module as an optional extension.

Experiments on four standard benchmark datasets (PEMS04, METR-LA, PEMS-BAY, and PEMS07) show that GC-MoE improves MAE over every individual expert and over a zero-parameter ensemble baseline, while MAPE is competitive and often improved, and the model trains only ${\sim}$17--18K parameters on top of ${\sim}$1.5M frozen backbone weights (roughly 1\% of total parameters). In addition, our ablation study indicates that a bounded output refinement layer can provide further gains, whereas node-adaptive ST-LoRA adapters were not beneficial in the routed multi-expert setting.

Several directions remain for future work. First, the current framework uses three expert architectures; incorporating a more diverse pool of experts may further increase complementary coverage across different traffic regimes. Second, our router operates at the node level with a shared set of global experts; a \emph{hierarchical} routing scheme that first clusters nodes into traffic regions and then routes within each region could improve scalability to very large networks while capturing meso-scale spatial structure. Third, extending the static pathway with \emph{learnable} structural or positional encodings derived from random-walk kernels may capture richer graph-structural information. Fourth, investigating \emph{cross-city transfer}, training the router on one city and deploying it on another with a different topology but similar traffic dynamics, would test the generalizability of graph-conditioned routing and move toward foundation-model paradigms for urban spatio-temporal prediction~\cite{yuan2024unist}. Fifth, an important next step is to evaluate whether the same routing principle remains effective on other spatio-temporal graph learning tasks and datasets with different graph semantics and temporal dynamics. Finally, a deeper analysis of routing dynamics over time (e.g., how expert preferences shift during peak vs.\ off-peak hours or incident conditions) could yield interpretable insights into the complementary roles learned by different ST-GNN architectures.

\bibliographystyle{IEEEtran}
\bibliography{references}

\end{document}